\pdfoutput=1

\documentclass[11pt]{article}

\usepackage[final]{acl}
\usepackage{booktabs} 
\usepackage{graphicx} 
\usepackage{times}
\usepackage{latexsym}
\usepackage{xcolor}
\usepackage[T1]{fontenc}
\usepackage[utf8]{inputenc}
\usepackage{natbib}
\usepackage{microtype}
\usepackage{subcaption}
\usepackage{amsfonts}
\usepackage{amsmath}
\usepackage{multirow}
\usepackage{multicol}
\usepackage{makecell}
\usepackage{adjustbox}
\usepackage{tabularx}
\usepackage{inconsolata}
\usepackage{placeins}

\title{DAPI: Domain Adaptive Toxicity Probe Vector Intervention 
for Fine-Grained Detoxification}

\author{Cho Hyeonsu$^{1}$\thanks{First author} \\
  SungKyunKwan University \\
  Republic of Korea \\
  \texttt{tacit3233@skku.edu} \\\And
  Dooyoung Kim$^{2}$\thanks{Second author} \\
  SungKyunKwan University \\
  Republic of Korea \\
  \texttt{kdysunleo98@gmail.com} \\\And
  Youngjoong Ko\thanks{Corresponding author} \\
  SungKyunKwan University \\
  Republic of Korea \\
  \texttt{yjko@skku.edu} \\}

\begin{document}
\maketitle
\begin{abstract}

There have been attempts to utilize linear probe for detoxification, with existing studies relying on a single toxicity probe vector to reduce toxicity. However, toxicity can be fine-grained into various subcategories, making it difficult to remove certain types of toxicity by using a single toxicity probe vector. To address this limitation, we propose a category-specific toxicity probe vector approach. First, we train multiple toxicity probe vectors for different toxicity categories. During generation, we dynamically select the most relevant toxicity probe vector based on the current context. Finally, the selected vector is dynamically scaled and subtracted from model. Our method successfully mitigated toxicity from categories that the single probe vector approach failed to detoxify. Experiments demonstrate that our approach achieves up to a 78.52\% reduction in toxicity on the evaluation dataset, while fluency remains nearly unchanged, with only a 0.052\% drop compared to the unsteered model.


\end{abstract}

\section{Introduction}
Pre-trained Language Models (LMs) possess strong language understanding and problem-solving capabilities. However, since these models are pre-trained on large-scale web datasets, they are likely to inherit bias and toxicity from their training data \cite{bias,rtp}. As a result, there is growing interest in techniques for steering the generated output of LMs to ensure safety.

 A linear probe refers to a classifier that uses a language model’s intermediate activations as input \cite{probe}, and it is originally developed for XAI. Recently, it has been utilized for steering LM generation in various aspects, such as toxicity \cite{dpo_toxic}, appropriateness \cite{appropriateness}, and truthfulness \cite{iti}. In detoxification, \citet{dpo_toxic} has demonstrated that intervening during the generation process effectively suppresses toxic outputs with the toxicity probe vector, which is obtained from toxic and non-toxic statements. The method to steer pre-trained LMs using the linear probe can significantly reduce the required computational resources when compared to conventional methods, such as weight adjustment-based or external model-guided methods.

\begin{figure}[t!]
    \centering
    \includegraphics[width=\linewidth]{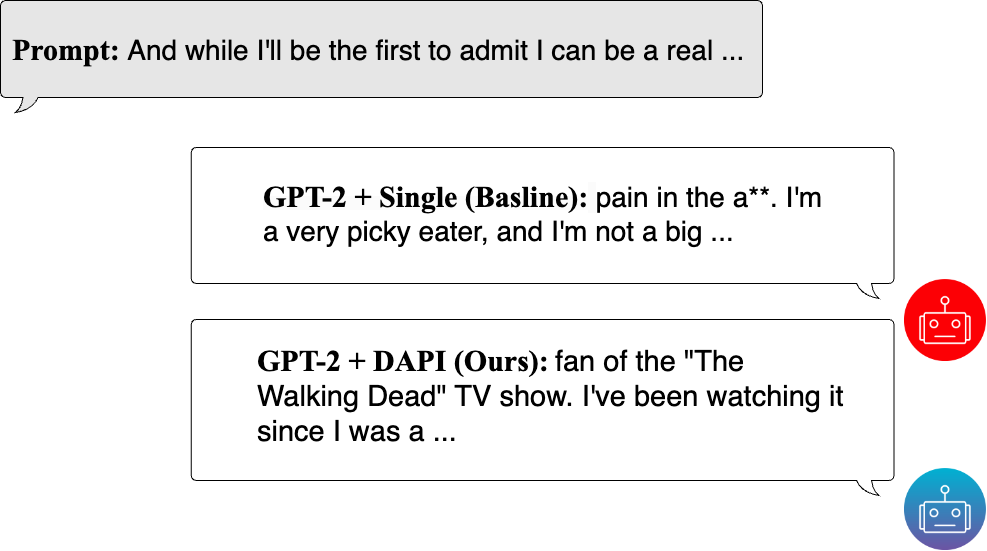}
    \caption{Continuations of the same non-toxic prompt from GPT-2 Large, generated with (red) single toxicity probe vector and with (blue) Domain-Adaptive Toxicity Probe Vector Intervention (DAPI). Without DAPI, the model produced toxic content despite the non-toxic prompt. In contrast, our method successfully prevented toxicity while maintaining fluency.
    \\ 
    \textcolor{red}{WARNING: THESE EXAMPLES ARE HIGHLY OFFENSIVE}}
    \label{fig:abstract}
\end{figure}

 Existing detoxification methods based on linear probe typically suppress toxic outputs using a single toxicity probe vector. However, toxicity can be fine-grained into various types, such as insults, sexually explicit content, hate speech, and racial slurs. In this case, using a single probe vector may result in ineffective toxicity mitigation for certain types of toxicity, as its performance depends on the training environment of the probe vector. As shown in Figure \ref{fig:abstract}, detoxification using a single probe vector struggles to effectively mitigate specific types of toxicity. We hypothesize that this issue arises due to dataset imbalance. Training a probe vector requires a labeled dataset, but publicly available datasets often have category imbalances, which can cause the over-representation of certain categories. As a result, while toxicity in majority categories can be successfully mitigated, detoxification for categories with a relatively smaller number of samples may fail. Furthermore, previous approach adjusts the intensity of toxicity suppression using a fixed intervention strength and this rigid adjustment may lead to unnecessary modifications even when the model is already safe, and it can ultimately degrade the fluency of the generated texts.
 
 To address these challenges, we proposes Domain Adaptive toxicity Probe vector Intervention (DAPI). First, multiple probe vectors are precomputed for each toxicity category through multi-class classification. During the generation process, the probe vector for intervention is dynamically selected based on its cosine similarity with the hidden states before the Feed-Forward Network (FFN) in the last Transformer decoder layer. The similarity is computed using the average hidden state across all token positions at each generation step. The selected probe vector is then dynamically scaled based on the steering state of the generation and subtracted from the last hidden state at last token in the last layer. Experiments demonstrate that our approach achieves up to 78.52\% toxicity reduction on the RealToxicityPrompts \cite{rtp} dataset, while fluency measured on the Wikitext-2 \cite{wikitext} dataset remained nearly unchanged, with only a 0.052\% drop compared to the unsteered model.

 To verify whether the distribution of the training dataset influenced the probe vector, we conducted a Category-wise experiment. Experimental results showed that using multiple probe vectors instead of a single probe vector improved toxicity reduction not only in majority categories but also in categories with relatively fewer samples. These findings demonstrate that using multiple probe vectors instead of a single one can prevent the probe vector from being biased toward a specific toxicity category. Furthermore, we conducted an analysis to assess whether the acquired multiple probe vectors capture distinctions between fine-grained toxicity categories. Additionally, through the ablation study on each component contribution, we observed that applying the dynamic scaling method instead of fixed scaling not only eliminated residual toxicity but also improved text quality.

 This paper makes the following three contributions:
 \begin{itemize}
     \item We propose DAPI to utilize multiple category-specific toxicity probe vectors for domain adaptive toxicity reduction
     
     \item We propose a cosine similarity regularization loss to acquire category-specific toxicity probe vectors.
     
     \item Experiments demonstrate that the proposed method outperforms all baseline models in toxicity reduction, while fluency remained nearly unchanged compared to the unsteered model.
 \end{itemize}

\begin{figure*}[t]
  \centering
  \includegraphics[width=\textwidth]{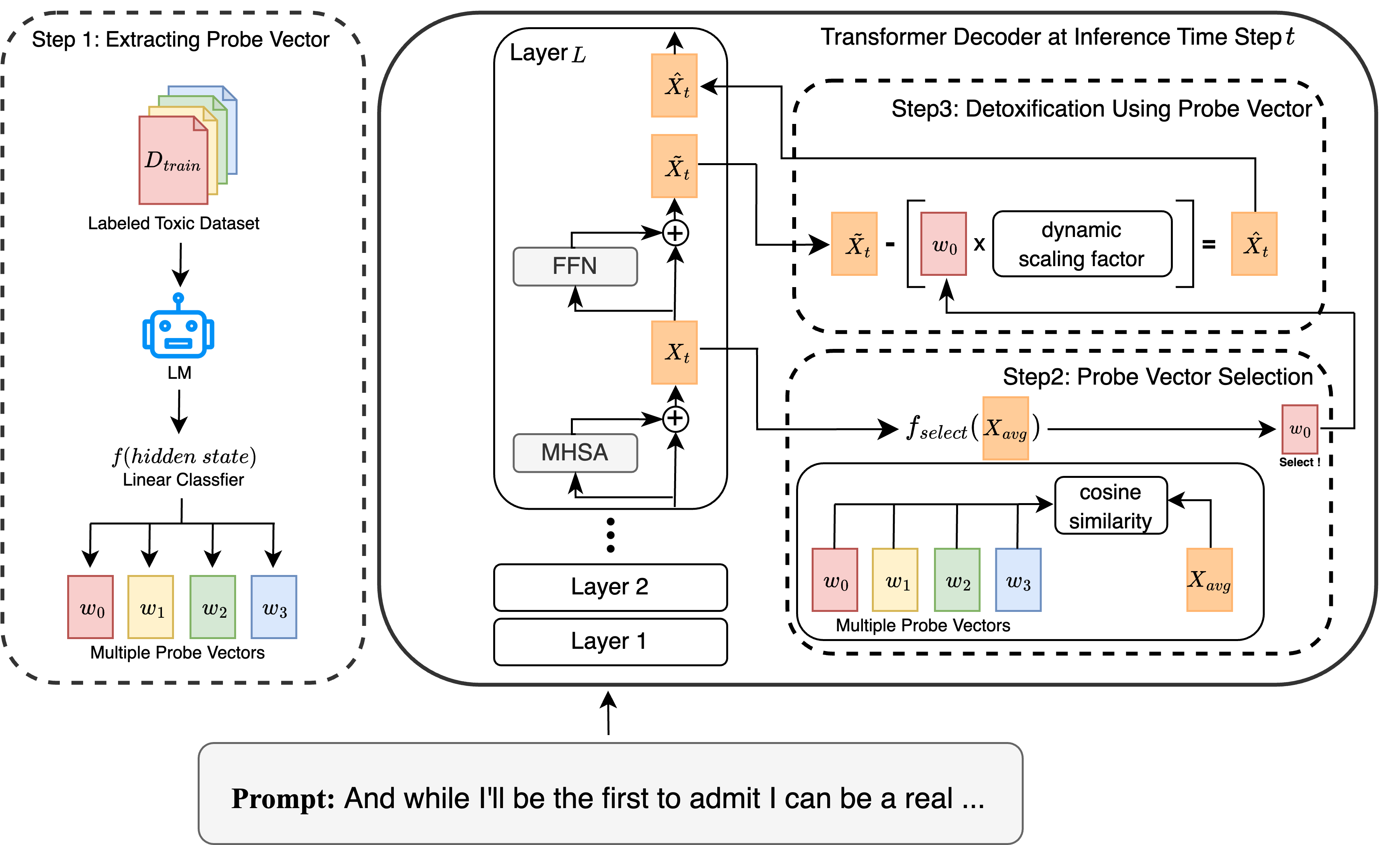}
  \caption{An overview of DAPI. It consists of three steps: \textbf{Step1: Extracting Probe Vector:} A linear classifier is trained using the last hidden state of the model to obtain probe vectors that represent distinct directional attributes for each toxicity category. \textbf{Step2: Probe Vector Selection:} At every time step \(t\) during inference, among the acquired probe vectors \(W\), the one most similar to the averaged hidden states before FFN, \(X_{avg}\) is selected. \textbf{Step3: Detoxification Using Probe Vector:} The selected probe vector is then scaled by a dynamic scaling factor and subtracted from the last hidden state $\tilde{X_t}$ in the model’s last layer. 
  }
  \label{fig:overview}
\vskip -.12in
\end{figure*}

\section{Related Work}

\paragraph{Linear Probe} The linear probe method involves training a linear layer on a specific intermediate activation from LM to extract directional vectors that encode desired attributes. This method identifies inherent attribute representations within the model and leverages them to guide controlled text generation. There are two main applications of this method. First application is \textbf{truthfulness control} \cite{iti}. To encourage pre-trained LMs to generate factually accurate responses, they analyze the differences in activation distributions between true and false statements. By training a binary classifier to distinguish between truth and falsehood, they obtain a linear probe. The accuracy of this linear probe is then used to identify which attention heads encode truthful information. This allows targeted interventions to adjust specific attention heads, ensuring the model produces more truthful outputs. Second application is \textbf{detoxification} \cite{dpo_toxic, surgery}. They trained a linear layer on a toxicity binary classification task using the language model’s intermediate activations as input, resulting in a toxicity probe vector that captures activation patterns associated with toxicity. Although their intervention method is differ, they all intervene in the model using this toxicity probe vector, to suppress toxic language in generated outputs without modifying the model’s weights.

\begin{table*}[t]
    \centering
    \begin{minipage}[t]{0.48\textwidth} 
        \centering
        \resizebox{\columnwidth}{!}{
        \begin{tabular}{llll}
        \toprule
        Category Pair & \multicolumn{2}{c}{Similarity (\%)} & Change (\%) \\
        \cmidrule(lr){2-3}
        & \textit{w/o cos-reg} & \textit{w cos-reg} & \\
        \midrule
        insult, identity      & 43.8     & 30.6  & \underline{-30.14\%}  \\
        insult, obscene       & 75.6     & 70.5  & \underline{-6.75\%}  \\
        insult, threat        & 25.1     & 2.0   & \underline{-92.03\%}  \\ 
        insult, other        & 45.8     & 44.1   & \underline{-3.71\%}  \\ 
        identity, obscene     & 32.4     & 21.1  & \underline{-34.88\%}  \\
        identity, threat      & 34.1     & 15.1  & \underline{-55.72\%}  \\
        identity, other      & 25.4     & 22.8  & \underline{-10.23\%}  \\
        obscene, threat       & 18.1     & 1.0   & \underline{-94.48\%}  \\
        obscene, other       & 43.2     & 43.4   & \underline{+0.46\%}  \\
        threat, other       & 15.4     & 0.8   & \underline{-94.81\%}  \\
        \bottomrule       
        \end{tabular}
        }
           \caption{
           Result of reduction in similarity between probe vectors after applying Cosine Similarity Regularization Loss. The "Change" column indicates the percentage change in similarity before and after applying regularization loss.  The term "identity" refers to the "identity-hate" category.
        }
        \label{tab:similarity_matrix}
    \end{minipage}
    \hfill 
    \begin{minipage}[t]{0.48\textwidth} 
        \centering
        \resizebox{\columnwidth}{!}{
        \begin{tabular}{ll}
        \hline
        \textbf{Probe Vector} & \textbf{Top Tokens} \\\cline{1-2}
        \hline 
        Single      & \textbf{f**k, a**hole, c**t, b**ch, s**t}, d**k          \\    \cline{1-2}    
        Multiple \textit{w/o cos-reg}                  \\  
        insult      & \textbf{f**k, a**hole, c**t, b**ch, s**t}, god**mn           \\  
        identity  & gays, homosexuals, homophobia, Jews             \\
        obscene      & \textbf{f**k, a**hole, c**t, b**ch, s**t}, wh**e                  \\
        threat      & murdering, decap, oreAnd, ThumbnailImage  \\ 
        other      & \textbf{f**k, a**hole, c**t, b**ch}, id***s, wh**e           \\ \cline{1-2}      
        Multiple \textit{w cos-reg}            \\  
        insult      & \textbf{f**k, b**ch}, dumb, Ni*, Stupid, vomit       \\  
        identity  & homosexuals, gays, lesbians, LGBT              \\
        obscene      & \textbf{f**k}, pu**y,\textbf{ a**hole}, scr*w, smoke, a**                 \\
        threat      & kill, stabbing, murdering, deaths, fatalities \\ 
        other      & \textbf{f**k}, c**t, disgusting, filthy, bastard          \\ \cline{1-2}
        \end{tabular}
        }
        \caption{Top tokens identified from probe vectors through vocabulary space projection, illustrating their alignment with predefined toxicity categories. The term "identity" refers to the "identity-hate" category. Tokens that appear redundantly are highlighted in bold.
        \\ 
        \textcolor{red}{WARNING: THESE EXAMPLES ARE HIGHLY OFFENSIVE}}
        \label{tab:top_token}
    \end{minipage}
\vskip -.15in
\end{table*}

\paragraph{Interpreting Value Vectors in Vocabulary Space} \label{sec:inter_value_vectors} 
Attempts to interpret the internal mechanisms of Transformer-based language models have been continuously explored \cite{keyvalue}. In particular, \citet{token} conducted an in-depth analysis of the role played by the FFN layers in the model’s prediction process. They interpreted the decoder weights of FFN as factors that either promote or suppress the likelihood of a specific tokens. \citet{dpo_toxic} leveraged this directional property by projecting the acquired probe vector onto the vocabulary space to identify which tokens be promoted. Inspired by this method, we aim to analyze how probe vectors for different toxicity categories would affect token generation and show they align with our predefined toxicity categories.

\section{Methodology}
Domain Adaptive toxicity Probe vector Intervention (DAPI) attempts to use the category-specific toxicity probe vectors to suppress toxic output.

\subsection{Extracting Probe Vector}
\label{sec:1}

\subsubsection{Multiple Toxicity Probe Vectors}
In this work, we utilize category-specific multiple toxicity probe vectors instead of a single toxicity probe vector. To extract probe vectors, we use the averaged last hidden states $\tilde{X}_{avg}$ as input. The classification task is then extended from binary toxicity classification to multi-class classification to distinguish between fine-grained toxicity categories. Thus, we leverage the four specific toxicity categories labeled in the dataset. The trained probe vectors achieved 76\% accuracy on the validation set.

\subsubsection{Other Category Toxicity Probe Vector}
The dataset for probe training contains sentences labeled as toxic that do not belong to any specific toxicity category. This subset consists of 5,707 sentences, which were used to train an other category toxicity probe vector. As a result, we utilize five category-specific probe vectors: (1) Insult, (2) Identity-Hate, (3) Obscene, (4) Threat, and (5) Other.

\subsubsection{Cosine Similarity Regularization Loss}
We needed to verify whether the acquired multiple probe vectors were effectively distinguished according to their respective toxicity categories. First we measured the cosine similarity between probe vectors. As shown in Table \ref{tab:similarity_matrix}, the average similarity between probe vectors was approximately 36\%, with the insult and obscene categories exhibiting particularly high similarity, close to 76\%. 
 
 Second, as mentioned in Section \ref{sec:inter_value_vectors}, we applied vocabulary space projection to further analyze the how probe vectors for different toxicity categories would affect token generation and show they align with our predefined toxicity categories. Following the method proposed by \citet{token}, we first extract the token embeddings from GPT-2’s embedding layer. Then, we compute the cosine similarity between the probe vector and all token embeddings. The top-k tokens with the highest similarity are decoded into actual text tokens using the tokenizer. This allowed us to visually inspect whether each probe vector captured semantically distinct directions. As shown in Table \ref{tab:top_token}, some probe vectors overlapped with specific word clusters, indicating that they primarily captured general toxicity rather than category-specific distinctions. Through the validation process above, we identified that the multi probe vectors exhibited high similarity and lacked sufficient differentiation across toxicity categories. To address this issue, we introduced Cosine Similarity Regularization Loss:

\begin{equation}
    \label{eq:cos_loss}
\mathcal{L}_{\text{R}} = \sum_{i=1}^{N} \sum_{j=i+1}^{N} \left| \frac{w_i}{\|w_i\|} \cdot \frac{w_j}{\|w_j\|} \right|
\end{equation}

\begin{equation}
    \label{eq:total_loss}
\mathcal{L}_{\text{total}} = \mathcal{L}_{\text{C}} + \lambda \mathcal{L}_{\text{R}}
\end{equation}

\noindent where \(\mathcal{L}_{\text{total}}\) is the total loss, \(\mathcal{L}_{C}\) and \(\mathcal{L}_{R}\) are the classification and regularization loss, respectively, and $\lambda \in \mathbb{R}$ is the regularization strength. \(\mathcal{L}_{R}\) is the Cross Entropy loss obtained with a linear layer on top of the last layer of language model as a multi-label classification head. \(\mathcal{L}_{R}\) is the Cosine Similarity Regularization Loss, which incorporated absolute sum of cosine similarities between multiple probe vectors. Let \(W=\{ w_1, w_2, \ldots, w_N \}\)  be the set of probe vectors, where each  \(w_i \in \mathbb{R}^d\)  represents the trained vector corresponding to one of the \(N\)  toxicity categories. \(N\) denotes the total number of categories, and each \(w_i\) is L2-normalized before computing the pairwise cosine similarity. Intuitively, a lower regularization loss value indicates that the probe vectors have learned more differentiated directional properties. The absolute sum of similarities was used to ensure that the common toxicity attribute remains intact, while enforcing distinctiveness across categories. Experimentally, the final classification accuracy of the multiple probe vectors reached 75\%.

\subsection{Probe Vector Selection}
\label{sec:2}
Based on the multi probe vectors trained in the previous stage, this step selects the most relevant probe vector for each token generation step. To achieve this, we compute the cosine similarity between each probe vector and the averaged hidden states before FFN, \(X_{avg}\) in the last layer. The probe vector with the highest similarity is then selected for the current generation step. However, if the highest similarity value is negative, it is considered that no toxicity needs to be removed at this step, and no probe vector is selected. Cases where a probe vector is selected even when the max similarity is negative can be observed in Table \ref{tab:neg_cos}.
\setlength{\abovedisplayskip}{0pt} 
\setlength{\belowdisplayskip}{0pt} 

\begin{equation}
\label{eq:cos}
\cos(w_i, X_{\text{avg}}) = 
\frac{w_i^\top X_{\text{avg}}}{\| w_i \|\| X_{\text{avg}} \|}, 
\quad \forall i \in \{1, \dots, N\}.
\end{equation}

\begin{equation}
\begin{aligned}
i^* &= \underset{i}{\operatorname{argmax}} \cos(w_i, X_{\text{avg}}), \\[6pt]
f_{select}(X_{\text{avg}}) &= 
\begin{cases}
w_{i^*}, & \text{if}~ \cos(w_{i^*}, X_{\text{avg}}) \ge 0,\\
\text{None}, & \text{otherwise}.
\end{cases}
\end{aligned}
\end{equation}

\vspace{10pt}
Additionally, instead of selecting only the probe vector with the max similarity, we also considered an approach that utilizes all category-specific probe vectors based on computed similarity. As shown in Table \ref{tab:selection}, we experimentally confirmed that selecting only the probe vector with the max similarity is more effective in toxicity reduction.
   	  
\subsection{Detoxification using Probe Vector}
\label{sec:3}
 This stage performs probe vector intervention by applying the probe vector selected in the previous stage for detoxification. The selected probe vector is scaled by a factor $\alpha$ and subtracted from the intermediate layer of the model. However, \citet{multi} has shown that applying additional vector interventions to already-controlled text results in diminishing improvements in toxicity reduction while potentially degrading fluency. To address this, we apply a dynamic scaling factor that adjusts at each token generation step, instead of using a fixed scaling factor.

 The existing dynamic scaling method computes the scaling factor for each generation step based on the KL-Divergence between the controlled and uncontrolled model’s probability distributions, which requires an additional forward pass. However, since our approach performs vector subtraction at the last hidden state $\tilde{X_t}$ from last layer instead of an intermediate layer, we can complete all computations within a single forward pass.

 This approach is not only intended to reduce computational overhead but is also more suitable because our probe vectors were trained using the hidden states from the last layer as input. Experimental results on layer-wise performance can be found in Appendix \ref{apd:layer}. This step consists of four stages. First, the probability distributions of the steered and unsteered models are computed using the LM head with the last-layer hidden state, with a fixed scaling factor of 12.5. Next, Top-p Sampling (nucleus sampling, \citealt{p}) is applied to each probability distribution, creating a vocabulary subset for each model, and their union is taken. Then, the probability distributions of both models are recomputed using the newly formed vocabulary subset. Finally, the KL-Divergence between the probability distributions is adjusted to match the desired scaling factor range, determining the optimal scaling value.

\begin{table*}[h]
    \centering
    \renewcommand{\arraystretch}{1.2}  
    \setlength{\tabcolsep}{5pt}  
    \begin{tabular}{lccccc}
        \toprule
        \textbf{Model} & \textbf{Toxicity (↓)} & \multicolumn{1}{c}{\textbf{Fluency (↓)}} & \multicolumn{3}{c}{\textbf{Diversity (↑)}} \\
        &  & Perplexity & Dist-1 & Dist-2 & Dist-3 \\
        \midrule
        GPT-2 & 0.4153 & 18.7715 & 0.8280 & 0.9256 & 0.9490 \\
        ActAdd & 0.3318 & 22.3115 & 0.7641 & 0.8766	& 0.9177 \\
        DExperts & \underline{0.1493} & 58.1782 & 0.8350 & 0.9181 & 0.9412 \\
        Single Probe Vector & 0.2241 & 20.1101 & 0.8450 & 0.9434 & 0.9655 \\
        \textbf{DAPI (ours)} & \textbf{0.0892} & 18.7813 & 0.8040 & 0.9096 & 0.9350 \\    
        \bottomrule
    \end{tabular}
    \caption{Main results of detoxifying generations using GPT-2 Large as the base LM. We evaluate toxicity, fluency, and diversity metrics. Lower toxicity and fluency (Perplexity) indicate better performance, while higher diversity (Dist-1, Dist-2, Dist-3) is preferred.}
    \label{tab:toxicity_results}
\end{table*}

\section{Experiments}

In this study, we validate the effectiveness of the proposed method through both automatic evaluation and human evaluation. Additionally, we conducted a category-wise evaluation to compare the effectiveness of using multiple probe vectors against a single probe vector in an imbalanced dataset environment. To assess the balance between toxicity reduction and generation quality, we design our experiments based on metrics commonly used in previous research \cite{token}. Moreover, distinct metrics were measured to evaluate the diversity of the model.

\subsection{Dataset}
\subsubsection{Probe Training Dataset}

 We use the Toxic Comment Classification Challenge dataset \cite{jigsaw}, released by Jigsaw on Kaggle, consists of Wikipedia discussion comments labeled by human evaluators for toxic behavior. It includes six columns: toxic, severe toxic, obscene, threat, insult, and identity attack, allowing multi-label classification. The dataset consists of 159,571 English comments, including 15,294 toxic comments and 144,277 non-toxic comments, with 478 threats, 7,877 insults, 8,449 obscene comments, and 1,405 identity attacks, reflecting an imbalanced distribution. We use a 90:10 split for training and validation.

\subsubsection{Evaluation Dataset}

\paragraph{REALTOXICITYPROMPTS}
To assess toxicity reduction performance, we utilize the “challenge” subset of the REALTOXICITYPROMPTS dataset \cite{rtp}. This subset consists of 1,199 prompts designed to elicit highly toxic outputs from language models. We evaluate the toxicity of generated text using the Perspective API\footnote{\url{https://github.com/conversationai/perspectiveapi}}, an automated toxicity detection tool. For each prompt, the model generates up to 20 tokens, and the Perspective API assigns a toxicity score to the generated output.
Additionally, we assess generation diversity using the REALTOXICITYPROMPTS dataset by computing the number of distinct n-grams in generated responses, scaled by the total number of generated tokens \cite{distinct}. Distinct-n quantifies generation diversity by computing the average number of unique n-grams, normalized by the total text length. We report Dist-1, Dist-2, and Dist-3, corresponding to unigram, bigram, and trigram diversity, respectively.

\paragraph{Wikitext}
To evaluate text generation quality, we measure Perplexity (PPL) using the Wikitext-2 dataset \cite{wikitext}. We compute sentence-level perplexity by iterating over the test set in fixed-length overlapping windows. A lower PPL value indicates better fluency, meaning the model generates more natural and coherent text.

\subsection{Baselines}
In all experiments, we use GPT-2 Large \cite{gpt2} as the base language model. Single Probe Vector \cite{dpo_toxic} utilizes a trained linear probe on a binary toxicity classification task. The input vector is obtained by averaging the last hidden states $\tilde{X}_{avg}$ of all token positions in a given sentence. That method also employs the Jigsaw dataset \cite{jigsaw}, split into 90:10 for training and validation. The trained single toxicity probe vector achieved 91\% accuracy on the validation set. We additionally compare our method with representative decoding time based, activation engineering. DExperts \cite{dexperts} is a decoding time based approach that combines a pretrained language model with toxic language model and non-toxic language model. ActAdd \cite{actadd} employs activation addition to steer language model outputs by utilizing the activation differences between pairs of input prompts.

\subsection{Setup}

 During multiple probe vector training, the weight for regularization loss is set to 0.01, and fixed scaling factor $\alpha$ is set to 17. When applying dynamic scaling, the average $\alpha$ value is measured at 16.8


\subsection{Automatic Evaluation}

\begin{figure*}[t]
  \centering
  \includegraphics[width=\textwidth]{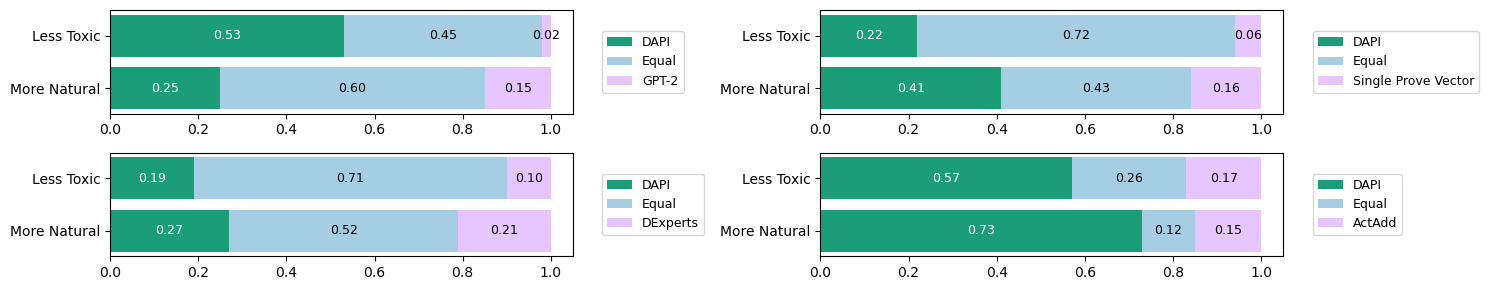}
  \caption{Results of human evaluation. 'Less toxic' means continuation is less toxic, 'More natural' means continuation is more natural and contextually coherent.}
  \label{fig:human}
\vskip -.12in
\end{figure*}

According to the automatic evaluation metrics in Table \ref{tab:toxicity_results}, our method (DAPI) outperforms all baseline methods in toxicity reduction while maintaining text generation quality and diversity. Compared to DExperts, DAPI achieved higher toxicity reduction, while significantly outperforming in terms of PPL. Moreover, DAPI requires training only a linear classifier, while DExperts necessitates fine-tuning two additional LMs, which makes our approach substantially more cost-efficient. These results indicate that for preserving text quality while detoxifying, utilizing a linear probe is more effective than a decoding-time-based approach. Compared to ActAdd, DAPI achieves superior performance in both toxicity reduction and text quality preservation. Furthermore, compared to Single Probe Vector, which also utilizes a linear probe, DAPI achieves better performance in both toxicity reduction and text quality. This result suggests that dynamically selecting category-specific probe vectors and adjusting the scaling factor based on the control state is more effective in preserving text quality while detoxifying, rather than applying a fixed scaling factor to a single probe vector.


\begin{table}[h]
    \centering
    \resizebox{\linewidth}{!}{  
    \begin{tabular}{lcccc}
        \toprule
        \multicolumn{5}{c}{\textbf{Toxicity (↓)}} \\
        & Obscene & Identity & Threat & Insult \\
        \midrule
        GPT-2 & 0.4552 & 0.4650 & 0.5336 & 0.3779 \\
        Single & 0.2355 & 0.3153 & 0.3130 & 0.2018 \\
        Multiple & \textbf{0.2125} & \textbf{0.1394} & \textbf{0.1481} & \textbf{0.1078} \\    
        \bottomrule
    \end{tabular}
    }
    \caption{Result of category-wise experiment. 'Single' means using single probe vector. And 'Multiple' means using multiple probe vectors without dynamic scaling. The term "Identity" refers to the "Identity-Hate" category.}
    \label{tab:category}
\end{table}

\subsection{Human Evaluation}
We conducted human evaluation on 100 random prompts sampling from the challenge subset of REALTOXICITYPROMPTS by four NLP expert evaluators. For each prompt, we compare four pairs of models: DAPI versus GPT-2 Large, DExperts, ActAdd, Single Toxicity Probe Vector. Each comparison pair is rated by four NLP expert evaluators, who select which of the two continuations is: (1) Less toxic: Continuation is less toxic, (2) More natural: Continuation is more natural and contextually coherent.
According to human evaluations, DAPI is rated as less toxic more often than all baselines \ref{fig:human}. In particular, it is rated equally natural compared to non-steeering model, yet less toxic.

\section{Category-wise Evaluation}

In the probe vector training environment, when using a dataset with category imbalances, we aimed to verify whether a single probe vector could successfully mitigate toxicity in majority categories while failing to detoxify categories with relatively fewer samples. To examine this, we additionally evaluated toxicity reduction performance for each category. For details on the Jigsaw dataset used for probe training, please refer to Appendix \ref{apd:category}.

Although REALTOXICITYPROMPTS does not have explicit category labels, each prompt is assigned category-specific toxicity scores measured by Perspective API. Based on this, we categorized each prompt according to the category with the highest toxicity score and conducted the evaluation accordingly. For this evaluation, we used the ‘threat’, ‘insult’, and ‘identity-attack’ categories as they are, while ‘profanity’ was grouped under ‘insult’, and ‘flirtation’ and ‘sexually-explicit’ were categorized as ‘obscene’. We also generates up to 20 tokens, and the Perspective API assigns a toxicity score to the generated output for each prompt. As shown in Table \ref{tab:category}, a single probe vector successfully mitigated toxicity in majority categories (Obscene, Insult) but failed to detoxify categories with relatively fewer samples (Identity, Threat). This demonstrates that a fine-grained approach to detoxification is more effective.

\begin{table*}[h]
    \centering
    \renewcommand{\arraystretch}{1.2}  
    \setlength{\tabcolsep}{5pt}  
    \begin{tabular}{lccccc}
        \toprule
        \textbf{Model} & \textbf{Toxicity (↓)} & \multicolumn{1}{c}{\textbf{Fluency (↓)}} & \multicolumn{3}{c}{\textbf{Diversity (↑)}} \\
        &  & Perplexity & Dist-1 & Dist-2 & Dist-3 \\
        \midrule
         GPT-2 & 0.4153 & 18.7715 & 0.8280 & 0.9256 & 0.9490 \\
        \textbf{DAPI} & \textbf{0.0892} & 18.7813 & 0.8040 & 0.9096 & 0.9350 \\   
        w/o dynamic scaling & 0.1398 & 18.7826 & 0.8141 & 0.9212 & 0.9466 \\ 
        w/o regularization loss & 0.1403 & 18.7826  & 0.8142 & 0.9211 & 0.9466 \\
        w/o other probe vector & 0.2742 & 18.7715 & 0.8301 & 0.9343 & 0.9581 \\
        \bottomrule
    \end{tabular}
    \caption{Results showing the contribution of each component. The table sequentially presents performance results starting from the base LM (GPT-2 Large), followed by the full method (DAPI), and then results without dynamic scaling, without regularization loss, and using the multiple probe vector without the other probe vector.}
    \label{tab:ablation}
\end{table*}

\begin{table}[t]
    \centering
    \resizebox{\linewidth}{!}{  
    \tiny
    \begin{tabular}{lcc}
        \toprule
        \textbf{Method} & \textbf{Toxicity (↓)} & \multicolumn{1}{c}{\textbf{Fluency (↓)}} \\
        &  & Perplexity \\
        \midrule    	  	 	 	
        Use Neg Cos & 0.1453 & 21.8967 \\
        \textbf{Pass Neg Cos} & \textbf{0.0892} & 18.7813 \\    
        \bottomrule
    \end{tabular}
    }
    \caption{Results of the negative cosine similarity experiment. ‘Use Neg Cos’ indicates that the probe vector is applied even when the highest cosine similarity is negative, while ‘Pass Neg Cos’ means that the case is skipped.}
    \label{tab:neg_cos}
\end{table}

\section{Ablation Study}

We first explore the influence of the probe vector with negative cosine similarity in Sec \ref{sec:neg}. Then, we analyze an alternative probe vector selection approach applying a weighted sum in Sec \ref{sec:select}. Finally, we examine the contribution of each component to detoxification performance in Sec \ref{sec:ablation}.

\subsection{Case of Negative Cosine Similarity}
\label{sec:neg}
As we stated in Section \ref{sec:2}, during the probe vector selection step, if the highest similarity value is negative, it is considered that no toxicity needs to be removed at this step. To validate this assumption, we conducted an experiment comparing cases where the probe vector is applied even when the highest cosine similarity is negative. As shown in Table \ref{tab:neg_cos}, applying the probe vector in such cases not only failed to effectively mitigate toxicity but also significantly degraded text quality. Based on these findings, we determined that in such cases, no probe vector should be selected.


\begin{table}[t]
    \centering
    \resizebox{\linewidth}{!}{ 
    \tiny
    \begin{tabular}{lcc}
        \toprule
        \textbf{Method} & \textbf{Toxicity (↓)} & \multicolumn{1}{c}{\textbf{Fluency (↓)}} \\
        &  & Perplexity  \\
        \midrule    	  	 	 	
        Weighted Sum & 0.1061 & 18.7992\\
        \textbf{Only Max} & \textbf{0.0892} & 18.7813  \\    
        \bottomrule
    \end{tabular}
    }
    \caption{Experimental results based on different probe vector selection methods. ‘Weighted Sum’ indicates a method where softmax is applied based on similarity scores, allowing the use of all probe vectors in a weighted sum. In contrast, ‘Only Max’ means that only the probe vector with the highest similarity is selected.}
    \label{tab:selection}
\end{table}

\subsection{Weighted Sum Selection Method}
\label{sec:select}

We experimented with an approach that utilizes all category-specific probe vectors by computing a weighted sum of multiple probe vectors based on their similarity scores instead of selecting only the probe vector with the max similarity (Table \ref{tab:selection}). We found that selecting only the probe vector with the max similarity is more effective in toxicity reduction and also shows slightly better performance in fluency and diversity.

\subsection{Each Component Contribution}
\label{sec:ablation}
  We also conducted experiments to analyze the contribution of each component in the proposed method to toxicity reduction performance (Table \ref{tab:ablation}). The significant performance difference with and without the other probe vector suggests that human-defined categories may not fully capture the complexity of toxicity, indicating the possibility of further fine-grained toxicity categorization.

While applying the regularization loss did not lead to a significant performance difference, as shown in Table \ref{tab:similarity_matrix}, the cosine similarity between probe vectors decreased after applying the regularization loss, reducing the average similarity from approximately 36\% to 25\%. Furthermore, as shown in Table \ref{tab:top_token}, each probe vector became better aligned with pre-defined toxicity categories and showed reduced overlap with specific word clusters. 

Additionally, the effectiveness of dynamic scaling in removing toxicity more efficiently suggests that a fixed scaling value was insufficient to fully mitigate toxicity in previous approaches. These results collectively demonstrate that each component of the proposed method contributes to both effective toxicity mitigation and text quality improvement.

\section{Conclusions}
We reveal that existing detoxification methods using a single steering vector fail to effectively mitigate certain categories of toxicity because they rely on a linear probe trained on an imbalanced dataset. To address the limitations of single probe vector approaches, we proposed Domain-Adaptive Toxic Probe Vector Intervention (DAPI), a three-stage framework that dynamically selects and adjusts category-specific probe vectors. Automatic evaluation experiments on the REALTOXICITYPROMPTS dataset demonstrated that DAPI reduces toxicity by up to 78.52\% and outperforms existing activation engineering and linear probe techniques in detoxification while maintaining generation quality, as indicated by Perplexity (PPL) and Distinct-n scores. This study highlights the effectiveness of fine-grained detoxification and suggests future research directions for extending this approach to other controlled text generation tasks.

\section*{Limitations}
While this study proposes and experimentally validates DAPI, it has several limitations that warrant further investigation. First, DAPI’s effectiveness beyond toxicity reduction remains unverified. Although our approach demonstrates success in detoxification, its applicability to other controllable text generation tasks, such as sentiment control, has not been explored. Additionally, applying the proposed method to other attributes requires a labeled dataset with category-specific annotations. Further analysis is needed to determine whether this method can be generalized to a wider range of attribute control tasks. Second, our experiments were conducted using GPT-2 Large, and we have not tested whether DAPI performs consistently in larger-scale LLMs. As model size increases, the impact of probe vector intervention may vary, requiring further investigation to assess its scalability and effectiveness in larger architectures. Third, while we hypothesized that the limitations of existing single steering vector approaches stem from imbalanced datasets and proposed a method to address this issue, we did not fundamentally resolve the dataset imbalance itself. Addressing these limitations in future research would enhance DAPI’s generalizability and enable its application to a broader range of controllable text generation tasks.

\section*{Ethics statement}
This study aims to improve safe and controllable text generation by developing Domain Adaptive toxic Probe vector Intervention for toxicity reduction. While our approach reduces harmful language in generated outputs, it relies on Perspective API for toxicity evaluation, which may reflect inherent biases in its training data. This could lead to inconsistent or unfair classifications, particularly for marginalized communities. Ensuring fairness in toxicity reduction remains an important challenge for future research. Although our primary focus is on the technical contributions, we recognize the potential risks associated with the misuse of our proposed intervention technique. In principle, the ability to manipulate activation space dynamically could be exploited to amplify harmful behaviors rather than mitigate them. While such risks exist, we believe that advancing research in controllable language generation remains essential, particularly in ensuring that model interventions remain transparent, adaptive, and aligned with ethical AI principles.

\bibliography{main}

\begin{thebibliography}{17}
\providecommand{\natexlab}[1]{#1}

\bibitem[{Alain(2016)}]{probe}
Guillaume Alain. 2016.
\newblock Understanding intermediate layers using linear classifier probes.
\newblock \emph{arXiv preprint arXiv:1610.01644}.

\bibitem[{cjadams et~al.(2017)cjadams, Sorensen, Elliott, Dixon, McDonald, nithum, and Cukierski}]{jigsaw}
cjadams, Jeffrey Sorensen, Julia Elliott, Lucas Dixon, Mark McDonald, nithum, and Will Cukierski. 2017.
\newblock Toxic comment classification challenge.
\newblock \url{https://kaggle.com/competitions/jigsaw-toxic-comment-classification-challenge}.
\newblock Kaggle.

\bibitem[{Gehman et~al.(2020)Gehman, Gururangan, Sap, Choi, and Smith}]{rtp}
Samuel Gehman, Suchin Gururangan, Maarten Sap, Yejin Choi, and Noah~A Smith. 2020.
\newblock Realtoxicityprompts: Evaluating neural toxic degeneration in language models.
\newblock \emph{arXiv preprint arXiv:2009.11462}.

\bibitem[{Geva et~al.(2022)Geva, Caciularu, Wang, and Goldberg}]{token}
Mor Geva, Avi Caciularu, Kevin~Ro Wang, and Yoav Goldberg. 2022.
\newblock Transformer feed-forward layers build predictions by promoting concepts in the vocabulary space.
\newblock \emph{arXiv preprint arXiv:2203.14680}.

\bibitem[{Geva et~al.(2020)Geva, Schuster, Berant, and Levy}]{keyvalue}
Mor Geva, Roei Schuster, Jonathan Berant, and Omer Levy. 2020.
\newblock Transformer feed-forward layers are key-value memories.
\newblock \emph{arXiv preprint arXiv:2012.14913}.

\bibitem[{Holtzman et~al.(2019)Holtzman, Buys, Du, Forbes, and Choi}]{p}
Ari Holtzman, Jan Buys, Li~Du, Maxwell Forbes, and Yejin Choi. 2019.
\newblock The curious case of neural text degeneration.
\newblock \emph{arXiv preprint arXiv:1904.09751}.

\bibitem[{Lee et~al.(2024)Lee, Bai, Pres, Wattenberg, Kummerfeld, and Mihalcea}]{dpo_toxic}
Andrew Lee, Xiaoyan Bai, Itamar Pres, Martin Wattenberg, Jonathan~K. Kummerfeld, and Rada Mihalcea. 2024.
\newblock A mechanistic understanding of alignment algorithms: A case study on dpo and toxicity.
\newblock \emph{ArXiv}, abs/2401.01967.

\bibitem[{Li et~al.(2015)Li, Galley, Brockett, Gao, and Dolan}]{distinct}
Jiwei Li, Michel Galley, Chris Brockett, Jianfeng Gao, and W.~Dolan. 2015.
\newblock A diversity-promoting objective function for neural conversation models.
\newblock \emph{ArXiv}, abs/1510.03055.

\bibitem[{Li et~al.(2023)Li, Patel, Vi'egas, Pfister, and Wattenberg}]{iti}
Kenneth Li, Oam Patel, Fernanda Vi'egas, H.~Pfister, and M.~Wattenberg. 2023.
\newblock Inference-time intervention: Eliciting truthful answers from a language model.
\newblock \emph{ArXiv}, abs/2306.03341.

\bibitem[{Liu et~al.(2021)Liu, Sap, Lu, Swayamdipta, Bhagavatula, Smith, and Choi}]{dexperts}
Alisa Liu, Maarten Sap, Ximing Lu, Swabha Swayamdipta, Chandra Bhagavatula, Noah~A Smith, and Yejin Choi. 2021.
\newblock Dexperts: Decoding-time controlled text generation with experts and anti-experts.
\newblock \emph{arXiv preprint arXiv:2105.03023}.

\bibitem[{Merity et~al.(2016)Merity, Xiong, Bradbury, and Socher}]{wikitext}
Stephen Merity, Caiming Xiong, James Bradbury, and Richard Socher. 2016.
\newblock Pointer sentinel mixture models.
\newblock \emph{arXiv preprint arXiv:1609.07843}.

\bibitem[{Radford et~al.(2019)Radford, Wu, Child, Luan, Amodei, Sutskever et~al.}]{gpt2}
Alec Radford, Jeffrey Wu, Rewon Child, David Luan, Dario Amodei, Ilya Sutskever, et~al. 2019.
\newblock Language models are unsupervised multitask learners.
\newblock \emph{OpenAI blog}, 1(8):9.

\bibitem[{Scalena et~al.(2024)Scalena, Sarti, and Nissim}]{multi}
Daniel Scalena, Gabriele Sarti, and Malvina Nissim. 2024.
\newblock Multi-property steering of large language models with dynamic activation composition.
\newblock \emph{arXiv preprint arXiv:2406.17563}.

\bibitem[{Sheng et~al.(2019)Sheng, Chang, Natarajan, and Peng}]{bias}
Emily Sheng, Kai-Wei Chang, Premkumar Natarajan, and Nanyun Peng. 2019.
\newblock The woman worked as a babysitter: On biases in language generation.
\newblock \emph{arXiv preprint arXiv:1909.01326}.

\bibitem[{Turner et~al.(2023)Turner, Thiergart, Leech, Udell, Vazquez, Mini, and MacDiarmid}]{actadd}
Alexander~Matt Turner, Lisa Thiergart, Gavin Leech, David Udell, Juan~J Vazquez, Ulisse Mini, and Monte MacDiarmid. 2023.
\newblock Activation addition: Steering language models without optimization.
\newblock \emph{arXiv e-prints}, pages arXiv--2308.

\bibitem[{von R{\"u}tte et~al.(2024)von R{\"u}tte, Anagnostidis, Bachmann, and Hofmann}]{appropriateness}
Dimitri von R{\"u}tte, Sotiris Anagnostidis, Gregor Bachmann, and Thomas Hofmann. 2024.
\newblock A language model's guide through latent space.
\newblock \emph{arXiv preprint arXiv:2402.14433}.

\bibitem[{Wang et~al.(2024)Wang, Yue, Lu, Shi, Zhao, Wang, Song, and Huang}]{surgery}
Huanqian Wang, Yang Yue, Rui Lu, Jingxin Shi, Andrew Zhao, Shenzhi Wang, Shiji Song, and Gao Huang. 2024.
\newblock Model surgery: Modulating llm's behavior via simple parameter editing.
\newblock In \emph{arXiv.org}.

\end{thebibliography}

\appendix

\section{Implementation Details}

For extracting probe vectors, we use GPT-2 Large as the frozen language model, while a linear classifier is trained on the training dataset. The probe model is trained for 20 epochs with a batch size of 128 and a learning rate of 5e-4. We apply weight decay of 0.01, set the warmup ratio to 0.1, and use a regularization loss weight of 0.01. For text generation, all methods utilize greedy search, generating a maximum of 20 tokens per sequence. All experiments are conducted on a single NVIDIA Quadro RTX 8000 GPU.

\section{Category Distribution of the Jigsaw Dataset \label{apd:category}}

The category distribution of the Jigsaw dataset, which was used for training the probe vectors, can be found in Table \ref{tab:toxicity_distribution}.
\begin{table}[h]
    \centering
    \begin{tabular}{lcc}
        \toprule
        Category & Count & Percentage (\%) \\
        \midrule
        Insult & 7,877 & 43.18 \\
        Obscene & 8,449 & 46.54 \\
        Identity-Attack & 1,405 & 7.65 \\
        Threat & 478 & 2.63 \\
        \bottomrule
    \end{tabular}
    \caption{Category distribution of the Jigsaw dataset.}
    \label{tab:toxicity_distribution}
\end{table}

\section{Layer-wise Ablation Study}

We conducted layer-wise experiments to determine the optimal layer position for performing the Detoxification using Probe Vector step. The results of these experiments can be found in Table \ref{tab:layer}.

\label{apd:layer}
\begin{table}[h]
    \centering
    \resizebox{\linewidth}{!}{ 
    \begin{tabular}{lccccc}
        \toprule
        \textbf{Layer} & \textbf{Toxicity (↓)} & \multicolumn{1}{c}{\textbf{Fluency (↓)}} & \multicolumn{3}{c}{\textbf{Diversity (↑)}} \\
        &  & Perplexity & Dist-1 & Dist-2 & Dist-3 \\
        \midrule    	    		 			 	 	
        1 layer (start) & 0.3561 & 18.7715 & 0.8304  & 0.9337 & 0.9587 \\
        12 layer (inter) & 0.3086   & 18.7715 & 0.8304  & 0.9337 & 0.9587 \\
        24 layer (inter) & 0.1639 & 18.7715 & 0.8304  & 0.9337 & 0.9587 \\
        \textbf{36 layer (last)} & \textbf{0.0892} & 18.7813 & 0.8040 & 0.9096 & 0.9350 \\    
        \bottomrule
    \end{tabular}
    }
    \caption{Results of layer-wise experiments.}
    \label{tab:layer}
\end{table}



\end{document}